\documentclass[letterpaper, 10 pt, conference]{ieeeconf}  % Comment this line out if you need a4paper

\IEEEoverridecommandlockouts                              % This command is only needed if 
\usepackage{verbatim}
\usepackage[pdftex]{graphicx}
\usepackage{epsfig}
\usepackage{amsmath,amssymb,amsfonts,bm,amscd} 
\usepackage{mathrsfs}
\usepackage{setspace}
\usepackage{fancyhdr}
\usepackage{algorithm} 
\usepackage{psfrag}
\usepackage{cite}
\usepackage[colorlinks=false]{hyperref}
\usepackage{color}
\usepackage[usenames,dvipsnames,svgnames,table]{xcolor}

\usepackage{hyperref}

\usepackage{pdfpages} 
\usepackage{epstopdf}
\usepackage{xcolor,colortbl}
\definecolor{grey1}{RGB}{192,192,192}
\definecolor{grey2}{RGB}{178,178,178}
\definecolor{grey3}{RGB}{150,150,150}
\definecolor{grey4}{RGB}{119,119,119}
\definecolor{grey5}{RGB}{77,77,77}
\definecolor{green1}{RGB}{112,173,71}
\definecolor{blue1}{RGB}{68,115,196}
\definecolor{red1}{RGB}{192,0,0}
\definecolor{yellow1}{RGB}{255,192,0}

\newcommand{\comRemove}{\textcolor{black}} % grey3
\newcommand{\comReview}{\textcolor{black}} % blue

\usepackage{subcaption} % subfigure
\let\labelindent\relax
\usepackage{enumitem}
\setlist[itemize,1]{leftmargin=\dimexpr 26pt}

\makeatletter
\def\namedlabel#1#2{\begingroup
    #2%
    \def\@currentlabel{#2}%
    \phantomsection\label{#1}\endgroup
}
\makeatother

\usepackage{flushend}

\title{\LARGE \bf
Knowledge Transfer Between Robots with Similar Dynamics for High-Accuracy Impromptu Trajectory Tracking
}

\author{Siqi Zhou$^{1}$, Andriy Sarabakha$^{2}$, Erdal Kayacan$^{3}$, Mohamed K. Helwa$^{1}$, and Angela P. Schoellig$^{1}$% <-this % stops a space
\thanks{$^{1}$Siqi Zhou, Mohamed K. Helwa, and Angela P. Schoellig are with the Dynamic Systems Lab (\href{http://www.dynsyslab.org}{http://www.dynsyslab.org}), Institute for Aerospace Studies, University of Toronto, Canada. The authors are also affiliated with the Vector Institute for Artificial Intelligence, Toronto. Mohamed K. Helwa is also with the Electrical Power and Machines Department, Cairo University, Egypt. Emails: siqi.zhou@robotics.utias.utoronto.ca, mohamed.helwa@robotics.utias.utoronto.ca,  schoellig@utias.utoronto.ca}%
\thanks{$^{2}$Andriy Sarabakha is with the School of Mechanical and Aerospace Engineering, Nanyang Technological University, Singapore. Email: andriy001@e.ntu.edu.sg}
\thanks{$^{3}$Erdal Kayacan is with the Department of Engineering, Aarhus University, Denmark. Email: erdal@eng.au.dk}
}

\begin{document}

\maketitle
\thispagestyle{empty}
\pagestyle{empty}

%%%%%%%%%%%%%%%%%%%%%%%%%%%%%%%%%%%%%%%%%%%%%%%%%%%%%%%%%%%%%%%%%%%%%%%%%%%%%%%%
\begin{abstract}

In this paper, we propose an online learning approach that enables the inverse dynamics model learned for a source robot to be transferred to a target robot  (e.g., from one quadrotor to another quadrotor with different mass or aerodynamic properties). The goal is to leverage knowledge from the source robot such that the target robot achieves high-accuracy trajectory tracking on arbitrary trajectories \textit{from the first attempt} with minimal data recollection and training. Most existing approaches for multi-robot knowledge transfer are based on post-analysis of datasets collected from both robots. In this work, we study the feasibility of \textit{impromptu} transfer of models across robots by learning an error prediction module online. In particular, we analytically derive the form of the mapping to be learned by the online module for exact tracking, propose an approach for characterizing similarity between robots, and use these results to analyze the stability of the overall system. The proposed approach is illustrated in simulation and verified experimentally on two different quadrotors performing impromptu trajectory tracking tasks, where the quadrotors are required to accurately track arbitrary hand-drawn trajectories from the first attempt. 

\end{abstract}

%%%%%%%%%%%%%%%%%%%%%%%%%%%%%%%%%%%%%%%%%%%%%%%%%%%%%%%%%%%%%%%%%%%%%%%%%%%%%%%%

\section{Introduction}
Machine learning techniques have been applied to many robot control problems with the goal of achieving high performance in the presence of uncertainties in the dynamics and the environment~\cite{nguyen2011model}. Due to the cost associated with data collection and training, approaches such as manifold alignment~\cite{Ammar2015AAAI,Bocsi2013IJCNN,Helwa2017IROS} and learning invariant features~\cite{Gupta2017ARXIV,daftry2016learning} have been proposed to transfer knowledge between robots and thereby increase the efficiency of robot learning. In these approaches, datasets on a set of sample tasks are initially collected from both robots. They are then used for finding a mapping offline to transfer knowledge from a source robot to a target robot. This transferred knowledge is expected to speed up the training of the target robot and enhance its performance in untrained tasks~\cite{Taylor2009JMLR}.

\begin{figure}[!t]
\centering
\includegraphics[width=\columnwidth]{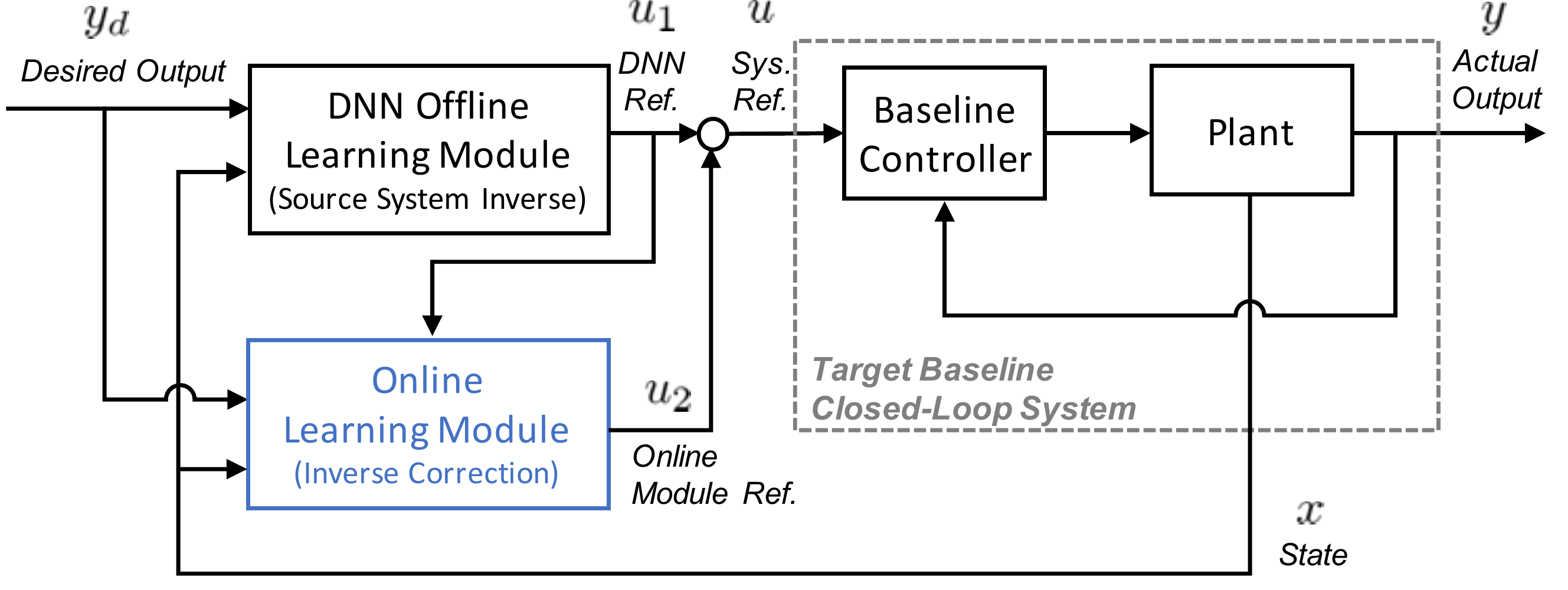}
\caption{Block diagram of the DNN-enhanced control architecture with online learning. The DNN module represents the inverse dynamics of a source system and is previously trained offline with a sufficiently rich dataset. During the testing phase, the DNN module is leveraged to enhance the tracking performance of a target system that shares some dynamic similarities with the source system. An online learning module (trained based on small sets of real-time data) further adjusts the reference generated by the DNN module to allow the target system to  achieve high-accuracy tracking on arbitrary trajectories from the first attempt (i.e., impromptu tracking). A video of this work can be found here: \href{http://tiny.cc/dnnTransfer}{http://tiny.cc/dnnTransfer}}
\label{fig:blkdiag}
\vspace{-1.8em}
\end{figure}

In this paper, we consider the problem of impromptu trajectory tracking, in which robots are required to track arbitrary trajectories accurately from the first attempt~\cite{li-icra17}. \comReview{Model-based techniques such as model predictive control (MPC) or the linear-quadratic regulator (LQR) can be used to solve tracking problems; 
%Techniques such as model predictive control (MPC) or PID control can be used to realize tracking functionalities; 
however, applying these techniques to achieve high tracking performance can be difficult as they rely on sufficiently accurate dynamics models or can be time-consuming to tune. In~\cite{li-icra17,zhou-cdc17}, we proposed a deep neural network (DNN)-based approach to enhance the tracking performance of black-box robot control systems. In particular, we showed that we can effectively enhance the tracking performance of a robot by training a DNN inverse dynamics module offline and then pre-cascading the module to the baseline system at test time.} For example, on 30 arbitrary, unseen hand-drawn trajectories, the DNN-enhancement approach reduced the tracking error of a quadrotor by an average of 43\%~\cite{li-icra17}. 

Motivated by recent work in transfer learning, in this work, we study the feasibility of leveraging the DNN model trained on one robot to enhance the performance of another robot in impromptu tracking tasks. In contrast to the existing approaches, where transfer mappings are usually found offline (e.g., \cite{Bocsi2013IJCNN,Gupta2017ARXIV}), we propose an online learning approach (Fig.~\ref{fig:blkdiag}) that allows a target robot using the DNN module from a source robot to achieve high-accuracy tracking \textit{impromptu} --- i.e., without additional data collection and training on sample tasks. With the online learning approach, we aim to significantly reduce the data recollection and training time usually required for enhancing the target robot performance. In this work, we
\begin{enumerate}[label={(\arabic*)}]
\item analytically derive the form of the mapping for the online module that allows the target system to achieve exact tracking,
\item present first results on characterizing system similarity between source and target systems and how it relates to the stability of the proposed overall learning system given modeling uncertainties, and
\item verify the effectiveness of the proposed approach in simulation and impromptu trajectory tracking experiments on quadrotors.
\end{enumerate}

The paper is organized as follows: We start with a brief review of the related work (Sec.~\ref{sec:related_work}) and provide some background on offline inverse learning (Sec.~\ref{sec:background}). Then, we formulate the online learning problem (Sec.~\ref{sec:problem}), discuss theoretical results (Sec.~\ref{sec:theory}), and illustrate the approach in simulation (Sec.~\ref{sec:simulation}) and in quadrotor experiments (Sec.~\ref{sec:experiments}). We conclude with a summary of the main results (Sec.~\ref{sec:conclusions}).
\section{Related Work}
\label{sec:related_work}
The problem of knowledge transfer or transfer learning has been studied in different application domains (e.g., natural language processing~\cite{blitzer2006domain}, computer vision~\cite{Wang2008Book}, and robot control~\cite{Bocsi2013IJCNN}). The common goal is to leverage existing data to accelerate and improve subsequent learning processes such that the costs (and potential risks) associated with data recollection can be reduced~\cite{Pan2010TKDE,Taylor2009JMLR}. In robotics, two directions of knowledge transfer have been considered: \textit{(i)} transfer across tasks and \textit{(ii)} transfer across robots. The former typically considers the transfer of knowledge from a source task to a target task to be performed by a single robot (e.g., \cite{fu2015one,pereida-acsp18,hamer2013knowledge}), while the latter considers the transfer of knowledge from a source robot to a target robot (e.g., \cite{Ammar2015AAAI,Bocsi2013IJCNN,pereida-ral18,Helwa2017IROS,Gupta2017ARXIV,daftry2016learning}). In this paper, we will focus on the latter. We aim to transfer the inverse dynamics model trained on one robot to enhance the tracking performance of another robot. The transferred inverse dynamics model is expected to generalize to arbitrary trajectories~\cite{li-icra17,zhou-cdc17}.

In the robot learning literature, and especially in reinforcement learning (RL), different approaches have been proposed to address the problem of knowledge transfer across different robots or domains. One of the approaches for cross-domain transfer is manifold alignment, where data from the source and target systems are collected for a set of sample tasks and are mapped to corresponding feature spaces (e.g., through dimensionality reduction) from which a transformation mapping between the source and target systems is found. This offline mapping can then be used to translate the policies trained on the source robot to the policies for the target robot~\cite{Ammar2015AAAI}, or map the data collected on the source robot to the target robot for model learning~\cite{Bocsi2013IJCNN}. Extension hereto~\cite{pereida-ral18,Helwa2017IROS} derive an optimal mapping for data transfer across robots from a control theory perspective. Other related work aims to learn and exploit a common feature space between the source and target robots while performing similar tasks~\cite{Gupta2017ARXIV,daftry2016learning}. In~\cite{daftry2016learning}, it is shown that the approach can effectively transfer control policies across different quadrotor platforms for autonomous navigation.

\textcolor{black}{In addition to the above, there are a few other lines of relevant work involving knowledge transfer. One of them is sim-to-real~\cite{marco-icra17,peng2018sim},  where the low-cost data from a simulation is exploited for accelerating the training on physical robots. Moreover, in meta learning, the learning parameters are optimized for initializing subsequent learning~\cite{finn2017one}. In~\cite{devin2017learning}, modularity in learning has also been proposed to maximize the utility of learned models.} 

Although recent literature demonstrates the possibility of transferring knowledge across robots, we address two additional aspects in our paper. The first aspect is \textit{impromptu} knowledge transfer without a-priori data collection on target systems. The second aspect is the impact of dynamic system similarity on the feasibility of knowledge transfer. An open question in the transfer learning literature is the issue of negative transfer (i.e., when the transfer adversely affects the target system)~\cite{Taylor2009JMLR}. While researchers have investigated task similarity in the context of task transfer problems~\cite{lazaric2012transfer}, discussions on system similarity for transferring knowledge across robots are rare. In this paper, we present theoretical results that associate system similarity to the feasibility of knowledge transfer across robots. 
\section{Background on Offline Inverse Learning}
\label{sec:background}
In this section, we provide more information about the DNN module (Fig.~\ref{fig:blkdiag}) to facilitate the  discussions in the following sections. In~\cite{zhou-cdc17}, we considered a nonlinear closed-loop baseline system represented by
\begin{equation}
\label{eqn:nonlinearSystem}
\begin{aligned}
x(k+1) &= f\left(x(k)\right) + g\left(x(k)\right)u(k)\\
y(k) &= h\left(x(k)\right),
\end{aligned}
\end{equation}
where $k\in\mathbb{Z}_{\ge 0}$ is the discrete time index, $x\in\mathbb{R}^n$ is the state of the system, $u\in\mathbb{R}$ and $y\in\mathbb{R}$ are the input and output of the system, respectively, and $f(\cdot)$, $g(\cdot)$, and $h(\cdot)$ are smooth functions.  % Following the convention in~\cite{sun2001analysis,jang1994iterative}, we denote $h \circ f$ as the composite of functions $h$ and $f$, and denote $f^i$ as the $i$-th composition of function $f$ with $f^0(x(t)) = x(t)$ and $f^i(x(t))=f^{i-1}\circ(f(x(t)))$. 
System~\eqref{eqn:nonlinearSystem} is said to have a relative degree $r$ at a point $(x_0,u_0)$ if $\frac{\partial}{\partial u}h \circ f^{p}(f(x(k))+g(x(k))u(k))= 0$ for $p=0,...,r-2$ for all points $(x,u)$ in the neighbourhood of $(x_0,u_0)$, and $\frac{\partial}{\partial u}h \circ f^{r-1}\big(f(x(k))+g(x(k))u(k)\big)\neq 0$ at $(x_0,u_0)$, where $(h \circ f)(x)$ is $h(f(x))$, and $f^i$ is the $i$-th composition of function $f$ with $f^0(x(t)) = x(t)$ and $f^i(x(t))=f^{i-1}\circ(f(x(t)))$~\cite{jang1994iterative}. For a system with a relative degree $r$, one may relate its input and output by $y(k+r) = h\circ f^{r-1}\big(f(x(k))+g(x(k))u(k)\big)$. 
%\begin{equation}
%y(k+r) = h\circ f^{r-1}\big(f(x(k))+g(x(k))u(k)\big).
%\end{equation}
For many practical systems (e.g., manipulators), the output $y(k+r)$ can be written as an affine function in the input $u(k)$:
\begin{equation}
\label{eqn:inputOuputNonlinear}
y(k+r)=\mathcal{F}\left(x(k)\right) + \mathcal{G}\left(x(k)\right)u(k),
\end{equation}
where $\mathcal{F}\left(x(k)\right) = h\circ f^r(x(k))$ and $\mathcal{G}\left(x(k)\right) = \frac{\partial}{\partial u}h \circ f^{r-1}(f(x(k))+g(x(k))u(k))$~\cite{jang1994iterative,sun2001analysis}. From Eqn.~\eqref{eqn:inputOuputNonlinear}, 
% \comReview{assuming $\mathcal{G}(x(k))\neq 0$ for all $x$ in the neighbourhood of~$x_0$}, 
one can show that the \comReview{reference signal} for exact tracking (i.e., $y(k+r)=y_d(k+r)$) is %$u(k) = \frac{1}{\mathcal{G}\left(x(k)\right)}\left(y_d(k+r) - \mathcal{F}\left(x(k)\right)\right)$. 
\begin{equation}
\label{eqn:controlLawNonlinear}
u(k) = \frac{1}{\mathcal{G}\left(x(k)\right)}\left(y_d(k+r) - \mathcal{F}\left(x(k)\right)\right).
\end{equation}
%By examining Eqn.~\eqref{eqn:controlLawNonlinear}, the optimal $u(k)$ for achieving exact tracking is a function of current state $x(k)$ and desired output $y_d(k+r)$, where $r$ is the relative degree of the baseline system. 
In more general cases, we can assume that the reference $u(k)$ for exact tracking is a nonlinear function of the state $x(k)$ and the future desired output $y_d(k+r)$. 
%\begin{equation}
%\label{eqn:controlLawNonlinear_General}
%u(k)=F_\text{nn}(x(k),y_d(k+r)),
%\end{equation}
%where $F_\text{nn}$ is a generic nonlinear function. 
In \cite{li-icra17,zhou-cdc17}, we showed that, for an unknown, minimum-phase, nonlinear baseline system with a well-defined relative degree, we can train a DNN module that approximates \comReview{the closed-loop inverse dynamics in Eqn.~\eqref{eqn:controlLawNonlinear} and effectively enhances the tracking performance of the baseline system}. In particular, in the training phase of the DNN module, we construct a dataset with input $\{x(k),y(k+r)\}$ and output $\{u(k)\}$ based on the input-output response data from the baseline system. In the testing phase, the DNN module is pre-cascaded to the baseline system to adjust the reference $u(k)$ based on the current state $x(k)$ and desired output $y_d(k+r)$~(see Fig.~\ref{fig:blkdiag}). %In~\cite{li-icra17}, it is experimentally shown on quadrotor vehicles that the offline trained DNN is able to improve the tracking performance on 30 arbitrary hand-drawn trajectories by 43\% on average.

Although we considered stable closed-loop baseline systems in~\cite{zhou-cdc17}, the results can be extended to that for stablizable open-loop plants. The approach in~\cite{zhou-cdc17} decouples the problem of stabilization from the problem of improving tracking performance, which makes the overall learning-based approach more effective and less prone to instabilities.

\section{Problem Formulation}
\label{sec:problem}
\comRemove{We consider the control architecture in Fig.~\ref{fig:blkdiag} and study the knowledge transfer problem that allows the DNN module trained on a source robot system to enhance the impromptu tracking performance of a target robot system that has different dynamics.} As in~\cite{zhou-cdc17}, the source and target robot systems are closed-loop systems whose dynamics can be represented by Eqns.~\eqref{eqn:nonlinearSystem} and~\eqref{eqn:inputOuputNonlinear}. 
We assume that:
\begin{itemize}
    \item[\text{\namedlabel{assumption:stability}{(A1)}}] The source and target systems are input-to-state stable~\cite{sontag2008input}.
    \item[\text{\namedlabel{assumption:inverse}{(A2)}}] The source and target systems \textit{(i)} have well-defined and the same relative degree, and \textit{(ii)} are minimum phase. 
    \item[\text{\namedlabel{assumption:desiredTrajectory}{(A3)}}] The desired trajectory $y_d$ is bounded, and a preview of $y_d(k+r)$ is available at time step $k$.
\end{itemize}
Note that \ref{assumption:stability} and \ref{assumption:inverse} are necessary for safe operations and for applying the DNN inverse learning~\cite{zhou-cdc17}. In \ref{assumption:inverse}, we also assume that the source and target systems have the same relative degree to simplify the analysis. \textcolor{black}{This condition holds, for instance, if the two robots have similar structures but different parameters (e.g., masses and dimensions).} For \ref{assumption:desiredTrajectory}, the relative degree of a system is typically a small integer bounded by the system order, and a preview of $r$ time steps of the desired trajectory can typically be achieved by online and offline trajectory generation algorithms.

\section{Theoretical Results}
\label{sec:theory}
In this section, we consider the control architecture in Fig.~\ref{fig:blkdiag} and provide theoretical results related to the knowledge transfer problem. We denote $u_1$ as the reference from the DNN module trained on the source system and $u_2$ as the reference from the online learning module. The overall reference to the target baseline system $u(k)$ is given by
\begin{equation}
\label{eqn:referenceCommand}
u(k) = u_1(k) + u_2(k).
\end{equation}
Below we derive an expression of $u_2(k)$ for achieving exact tracking in Sec.~\ref{subsec:optimalMapping}, propose a characterization of system similarity in Sec.~\ref{subsec:similarity}, and analyze the stability of the overall system in the presence of uncertainties in Sec.~\ref{subsec:stability}.

\subsection{Reference Adaptation for Exact Tracking}
\label{subsec:optimalMapping}
In this subsection, we derive an expression for $u_2(k)$ such that $u(k)$ achieves exact tracking $y(k+r) = y_d(k+r)$, where $y$ and $y_d$ are the desired and actual outputs of the target system, and $r$ is the system relative degree.

A common approach for high-accuracy trajectory tracking is to adapt the reference input of a nominal controller based on the observed tracking errors. For instance, in PD-type iterative learning control (ILC), proportional and derivative tracking error terms are added to the reference in each iteration to improve the tracking performance over a sequence of trials~\cite{andreas}. In distal teacher inverse dynamics learning, the tracking error is proposed as the cost function for updating the weights of a neural-network-based controller online to achieve improved tracking~\cite{jordan1992forward}. In this work, we similarly consider an online learning approach that adapts the reference of the DNN module $u_1(k)$ based on the tracking error. In particular, we justify below that the reference $u_2(k)$ can be approximated by
\begin{equation}
\label{eqn:adaptationForm}
u_2(k) = \alpha\:e_p(k+r),
\end{equation}
where $\alpha$ is an adaptation gain, and $e_p(k+r)$ is a prediction of the tracking error $r$ time steps ahead. 

We consider a nonlinear target system~\eqref{eqn:nonlinearSystem}, \eqref{eqn:inputOuputNonlinear}:
\begin{equation}
\label{eqn:targetForward}
y(k+r)=\mathcal{F}_t\left(x(k)\right) + \mathcal{G}_t\left(x(k)\right)u(k),
\end{equation}
where 
$\mathcal{F}_t\left(x(k)\right) = h_t\circ f_t^r(x(k))$ and $\mathcal{G}_t\left(x(k)\right) = \frac{\partial}{\partial u}h_t \circ f_t^{r-1}(f_t(x(k))+g_t(x(k))u(k))$, 
and $f_t(\cdot)$, $g_t(\cdot)$, and $h_t(\cdot)$ are the corresponding nonlinear functions in Eqn.~\eqref{eqn:nonlinearSystem}. In addition to the target system, we consider a source system, which the DNN module is trained on. This system is similarly represented in the form of Eqn.~\eqref{eqn:inputOuputNonlinear}. As discussed in~\cite{zhou-cdc17}, the underlying function approximated by the DNN is
\begin{equation}
\label{eqn:dnnMapping}
u_1(k) = \frac{1}{\mathcal{G}_s\left(x(k)\right)}\left(y_d(k+r) - \mathcal{F}_s\left(x(k)\right)\right),
\end{equation}
where $\mathcal{F}_s\left(x(k)\right)$ and $\mathcal{G}_s\left(x(k)\right)$ are defined analogously to those of the target system.  
By substituting Eqns.~\eqref{eqn:referenceCommand} and \eqref{eqn:dnnMapping} into Eqn.~\eqref{eqn:targetForward}, one can see that the ideal reference $u_2(k)$ for achieving exact tracking is
\begin{equation}
\label{eqn:optimalMapping}
u_2(k) = \alpha^*e_p^*(k+r),
\end{equation}
where $\alpha^* = \frac{1}{\mathcal{G}_t\left(x(k)\right)}$ and
\begin{equation}
    \begin{aligned}
    \label{eqn:optimalErrorPrediction}
    e_p^*(k+r) &= y_d(k+r) - \mathcal{F}_t\left(x(k)\right) -  \mathcal{G}_t\left(x(k)\right)u_1(k).
    % \\
    % &= y_d(k+r) - \mathcal{F}_t\left(x(k)\right) -\\
    % &\hspace{3em}\frac{\mathcal{G}_t\left(x(k)\right)}{\mathcal{G}_s\left(x(k)\right)}\left(y_d(k+r) - \mathcal{F}_s\left(x(k)\right)\right).
    \end{aligned}
\end{equation}

\vspace{0.9em}
\noindent\textbf{Insight~1. Ideal Mapping for Exact Tracking.} In order to achieve exact tracking, the online learning module should predict the tracking error of the target system that would result from applying $u_1(k)$. The predicted error is scaled by a gain $\alpha^*=\frac{1}{\mathcal{G}_t(x(k))}$, where $\mathcal{G}_t(x(k)) = \frac{\partial y(k+r)}{\partial u(k)}$.\\[-0.5em] %and can be intuitively thought of as a characterization of the aggressiveness of the target system.\\[0.5em]
\vspace{0.55em}

The error prediction in Eqn.~\eqref{eqn:optimalErrorPrediction} depends on the current state $x(k)$, the reference $u_1(k)$ from the DNN module, and the future desired output $y_d(k+r)$. When the dynamics of the source and the target systems are not known, one may use supervised learning to train a model online to approximate Eqn.~\eqref{eqn:optimalErrorPrediction}. We present a general approach for training this online model in Remark~1. 

\noindent\textbf{Remark~1. Online Learning for Error Prediction.} 
For training an online model to approximate Eqn.~\eqref{eqn:optimalErrorPrediction}, at each time step $k$, one may construct a dataset with paired inputs $\{x(p-r),u(p-r),y_d(p)\}$ and outputs $\{y_d(p)-y(p)\}$ over the past $N$ time steps $p = k-N, ..., k$. % where $N$ is the size of the dataset. 
The error $e_p(k+r)$ can then be predicted using the online model with input $\mathcal{I} = [x(k),u_1(k),y_d(k+r)]$.\\[-0.8em]

Given the predicted error $e_p(k+r)$, another component to be determined for computing $u_2(k)$ is the gain~$\alpha$. 
With an online model $F\left(x(k),u_1(k),y_d(k+r)\right)$ approximating Eqn.~\eqref{eqn:optimalErrorPrediction}, it can be shown that $\alpha^*$ can be obtained from $\hat{\alpha}^* = -\left(\partial F/\partial u_1\right)^{-1}$. In practice, due to noise in the systems, the online estimation of $\alpha^*$ can be non-trivial. In Sec.~\ref{subsec:stability}, we provide an analysis to examine the stability of the overall system when $\alpha^*$  is approximated by a constant and also when the estimation of $e_p^*(k+r)$ by the online model is inexact.

\subsection{System Similarity}
\label{subsec:similarity}
The concept of task similarity has been introduced in the RL literature to address the issue of negative knowledge transfer in task transfer learning problems~\cite{lazaric2012transfer}. In this subsection, we propose a characterization of system similarity for impromptu knowledge transfer problems, where an inverse module is transferred across two robot systems. 

We consider two systems are similar if at any given state $x(k)$, the application of an input $u(k)$ to the systems results in similar outputs $y(k+r)$~\cite{Whorton2008AIAA}. For the similarity discussion, we assume linear or linearized source and target systems to simplify our analysis:
\begin{equation}
\label{eqn:linearSystem}
\begin{aligned}
x(k+1) &= Ax(k) + Bu(k)\\
y(k) &= Cx(k),
\end{aligned}
\end{equation}
where $x\in\mathbb{R}^n$ is the state, $u\in\mathbb{R}$ is the input, $y\in\mathbb{R}$ is the output, and ($A$, $B$, $C$) are constant matrices. It can be shown that the input and output of system~\eqref{eqn:linearSystem} are related by
\begin{equation}
\label{eqn:inputOuputLinear}
y(k+r) = \mathcal{A}x(k)+\mathcal{B}u(k).\end{equation}
where $\mathcal{A} = CA^r$ and $\mathcal{B} = CA^{r-1}B$, and $r$ is the relative degree of system~\eqref{eqn:linearSystem}. From Eqn.~\eqref{eqn:inputOuputLinear}, the input-output relationship is fully characterized by $\mathcal{A} $ and $\mathcal{B}$, which can be thought as the state-to-output gain vector and the input-to-output gain, respectively. Based on the relationship in Eqn.~\eqref{eqn:inputOuputLinear}, we define a vector $S$ to characterize the similarity of the source and target systems:
\begin{align}
\label{eqn:similarityLinearS}
S = \begin{bmatrix}
S_1 & S_2
\end{bmatrix},
\end{align}
where $S_1 = 1-\frac{\mathcal{B}_t}{\mathcal{B}_s}$, $S_2 = \mathcal{A}_t - \frac{\mathcal{B}_t}{\mathcal{B}_s} \mathcal{A}_s$, and the subscripts $s$ and $t$ denote the source and the target system. The terms $S_1$ and $S_2$, respectively, characterize the differences in the input-to-output gain and state-to-output gain vector of the source and target systems. Note that $S = 0$ if and only if $\mathcal{A}_t=\mathcal{A}_s$ and $\mathcal{B}_t=\mathcal{B}_s$ (i.e., the state-to-output and input-to-output gains of the systems are identical). 

\color{black}
\subsection{Stability in the Presence of Uncertainties}
\label{subsec:stability}
In this subsection, we use the concept of system similarity and analyze the stability of the target system when the gain $\alpha^*$ is approximated by a constant $\alpha$ and the prediction of the future error $e_p^*(k+r)$ is not exact. We focus on system~\eqref{eqn:linearSystem} and make the following assumptions:
 \begin{itemize}
     \item[\text{\namedlabel{assumption:u1}{(A4)}}] The output of the offline DNN $u_1(k)$ corresponds to the inverse of the source system $u_1(k) = \frac{1}{\mathcal{B}_s} \left(y_d(k+r)-\mathcal{A}_sx(k)\right)$, where $\mathcal{A}_s$ and $\mathcal{B}_s$ are the gains of the source system, and $x(k)$ and $y_d(k+r)$ are the state and desired output of the target system.
     \item[\text{\namedlabel{assumption:errorUncertainty}{(A5)}}] The error in the prediction $\Lambda =  e_p^*(k+r)-e_p(k+r)$ can be bounded as follows: $\Lambda \le \beta_1 ||y_d(k+r)|| + \beta_2||x(k)|| + \beta_3$, where $\beta_1$, $\beta_2$, and $\beta_3$ are positive constants, and $||\cdot||$ is the Euclidean norm.
\end{itemize}
In addition, by (A1), the target system is input-to-state stable. It can be shown that the state of system~\eqref{eqn:linearSystem} can be bounded as follows: $||x||_{\infty}\le L_1||u||_{\infty} + L_2||x_0||$, where $||x||_{\infty} = \sup_k \{||x(k)||\}$, $||u||_{\infty} = \sup_k \{||u(k)||\}$, and $L_1$ and $L_2$ are positive constants.

\vspace{0.5em}
\noindent\textbf{Lemma~1. Stability.} Consider a target system represented by Eqn.~\eqref{eqn:linearSystem} and the control architecture in Fig.~\ref{fig:blkdiag}, where the reference of the online learning module $u_2(k)$ has the form of Eqn.~\eqref{eqn:adaptationForm}. Under \ref{assumption:stability}, \ref{assumption:u1}, and \ref{assumption:errorUncertainty}, the overall system is bounded-input-bounded-state (BIBS) stable if 
\begin{align}
\label{eqn:condition}
|\alpha|\left(|| S_2 || +\beta_2\right)<\frac{\beta_4}{L_1},
\end{align}
where $\beta_4 =1-L_1 \left\vert\left\vert\frac{\mathcal{A}_s}{\mathcal{B}_s} \right\vert\right\vert$. \\[0.5em]
\noindent\textit{Proof.} 
At a time step $k$, the output of the online learning module is $u_2(k) = \alpha\: e_p(k+r)$, where $\alpha$ is a constant gain and $e_p(k+r)$ is the predicted tracking error. 
%At a timestep $k$, 
The adjusted reference $u(k)$ sent to the target baseline system is $u(k) = u_1(k) + \alpha\: e_p(k+r)$, where $u_1(k) $ is the output of the offline DNN module. 
By \ref{assumption:u1} and \ref{assumption:errorUncertainty}, we can write $u(k)$ as
\begin{equation}
\label{eqn:u_k_intermediate}
    u(k) = \frac{1}{\mathcal{B}_s} \left(y_d(k+r)-\mathcal{A}_sx(k)\right) + \alpha\left(e^*_p(k+r) - \Lambda\right).
\end{equation}
For a target system represented by Eqn.~\eqref{eqn:linearSystem}, $e_p^*(k+r)$ in Eqn.~\eqref{eqn:optimalErrorPrediction} can be written as $e_p^*(k+r) = y_d(k+r) - \mathcal{A}_tx(k) - \mathcal{B}_tu_1(k)= y_d(k+r) - \mathcal{A}_tx(k) - \frac{\mathcal{B}_t}{\mathcal{B}_s}\left(y_d(k+r)-\mathcal{A}_sx(k)\right)$. By substituting the expression of $e_p^*(k+r)$ into Eqn.~\eqref{eqn:u_k_intermediate}, we obtain $u(k) = \left(\frac{1}{\mathcal{B}_s}+\alpha S_1 \right) y_d(k+r) -  \left(\frac{\mathcal{A}_s}{\mathcal{B}_s} + \alpha S_2 \right)x(k) - \alpha\Lambda$. 
Moreover, by \ref{assumption:stability} and \ref{assumption:errorUncertainty}, we can relate $||x||_\infty$ to $||y_d||_{\infty} = \sup_k \{||y_d(k)||\}$ by the following inequality:
\begin{equation}
\label{eqn:stateOutput}
\begin{aligned}
    ||x||_{\infty} &\le L_1\left(\left(\left\vert\frac{1}{\mathcal{B}_s}\right\vert+\left\vert\alpha\right\vert\left\vert S_1 \right\vert+\beta_1|\alpha| \right)||y_d||_\infty \right. \\ &\hspace{3.5em}\left.+\left(\left\vert\left\vert\frac{\mathcal{A}_s}{\mathcal{B}_s} \right\vert\right\vert+ \left\vert \alpha \right\vert\left\vert\left\vert S_2 \right\vert\right\vert +\beta_2|\alpha|\right) ||x||_\infty\right)\\
    &\hspace{3.5em}+L_1\beta_3|\alpha|+ L_2||x_0||.
\end{aligned}
\end{equation}
From Eqn.~\eqref{eqn:stateOutput}, if $1 - L_1\left(\left\vert\left\vert\frac{\mathcal{A}_s}{\mathcal{B}_s} \right\vert\right\vert+ \left\vert \alpha \right\vert\left\vert\left\vert S_2 \right\vert\right\vert +\beta_2|\alpha|\right) > 0$, or equivalently $|\alpha|\left(|| S_2 || +\beta_2\right)<\frac{\beta_4}{L_1}$, then the state of the system can be bounded as follows: $||x||_{\infty} \le \frac{L_1 \left(\left\vert\frac{1}{\mathcal{B}_s}\right\vert+\left\vert\alpha\right\vert\left\vert S_1 \right\vert+\beta_1|\alpha| \right)||y_d||_\infty +L_1\beta_3|\alpha|+ L_2||x_0||}{1 - L_1\left(\left\vert\left\vert\frac{\mathcal{A}_s}{\mathcal{B}_s} \right\vert\right\vert+ \left\vert \alpha \right\vert\left\vert\left\vert S_2 \right\vert\right\vert +\beta_2|\alpha|\right)}$. Now, if $y_d$ and hence $||y_d||_\infty$ are bounded, then the system state is bounded, and the overall system is BIBS stable.\hfill$\square$

Recall that, in Eqn.~\eqref{eqn:condition}, $\alpha$ is the gain of the online learning module, $S_2$ characterizes the similarity between the two systems, $\beta_1$ is associated with the uncertainty in the error prediction, and $L_1$ can be thought of as a characterization of the aggressiveness of the target system. The condition in Eqn.~\eqref{eqn:condition} can be interpreted for two scenarios: (\textit{i}) when $|\alpha|=0$  (i.e., the online module is inactive) and \textit{(ii)} when $|\alpha|\neq0$ (i.e., the online module is active). In scenario \textit{(i)}, the condition in Eqn.~\eqref{eqn:condition} reduces to $L_1  < \frac{1}{\left\vert\left\vert \mathcal{A}_s/\mathcal{B}_s\right\vert\right\vert}$, which can be interpreted as an upper bound on the relative aggressiveness of the source and target systems. When this condition is satisfied, the target system with the source system DNN module is stable. 
In scenario \textit{(ii)}, when the online learning module is active, the  condition in Eqn.~\eqref{eqn:condition} implies that if the source and target systems are more similar, that is $||S_2||$ is closer to 0, then there will be a greater margin for selecting $\alpha$ and higher tolerance for having uncertainties in the online prediction model. Moreover, based on the condition in Eqn.~\eqref{eqn:condition}, one may use probabilistic learning techniques to estimate the uncertainties in the predicted error $e_p(k+r)$ and calculate an upper bound on the magnitude of the fixed gain $\alpha$ for stability.\\[-0.5em]

\noindent\textcolor{black}{\textbf{Remark~2. Generalization to Nonlinear Systems.}} For nonlinear systems~\eqref{eqn:nonlinearSystem} with inputs and outputs related by Eqn.~\eqref{eqn:inputOuputNonlinear}, one can relate the outputs of the source and target systems by $y_t(k+r) = \vartheta_1\left(x(k)\right)y_s(k+r) + \vartheta_2\left(x(k)\right)$, where $\vartheta_1\left(x(k)\right)=\frac{\mathcal{G}_t\left(x(k)\right)}{\mathcal{\mathcal{G}}_s\left(x(k)\right)}$ and $\vartheta_2\left(x(k)\right)=\mathcal{F}_t\left(x(k)\right) -\frac{\mathcal{G}_t\left(x(k)\right)}{\mathcal{\mathcal{G}}_s\left(x(k)\right)}\mathcal{F}_s\left(x(k)\right)$. The relation between $y_t(k+r)$ and $y_s(k+r)$ can be used as an alternative for characterizing the similarity for the nonlinear systems. It is left for future work to perform a similar stability analysis for the nonlinear case.
\section{Simulation Illustration}
\label{sec:simulation}
In this section, we illustrate the proposed online learning approach with a simulation example. In~\cite{zhou-cdc17}, we considered a minimum phase closed-loop baseline system represented by
\begin{equation}
\label{eqn:system_sim_source}
\begin{aligned}
x(k+1) &= \left[\begin{matrix}
0 & 1\\-0.15 & 0.8
\end{matrix}\right]x(k)+\left[\begin{matrix}
0 \\ 1
\end{matrix}\right]u(k)\\
y(k) &= \left[\begin{matrix}
-0.2 & 1
\end{matrix}\right]x(k),
\end{aligned}
\end{equation}
and showed that a DNN module (Sec.~\ref{sec:background}) can be designed to enable the system to achieve exact tracking on untrained trajectories. In the following simulation study, we consider system~\eqref{eqn:system_sim_source} as the source system and leverage its offline DNN module to enhance the tracking performance of a target system that is represented by
\begin{equation}
\label{eqn:system_sim_target}
\begin{aligned}
x(k+1) &= \left[\begin{matrix}
0 & 1\\-0.24 & 1
\end{matrix}\right]x(k)+\left[\begin{matrix}
0 \\ 1
\end{matrix}\right]u(k)\\
y(k) &= \left[\begin{matrix}
-0.1 & 1
\end{matrix}\right]x(k). 
\end{aligned}
\end{equation}
Note that the source system~\eqref{eqn:system_sim_source} and the target system~\eqref{eqn:system_sim_target} are minimum phase and have relative degrees of 1. The source system has two poles at $\{0.3, 0.5\}$ and a zero at 0.2, while the target system has two poles at $\{0.4, 0.6\}$ and a zero at 0.1. 
When implementing the learning modules, we assume that the systems are black boxes, and we rely on only their input-output data and basic properties (e.g., relative degree).

\subsection{Learning Modules}
\label{subsec:simulationSetup}

\subsubsection{Offline Learning of Inverse Module}
\label{subsubsec:sim_offlineLearningModule}
The offline inverse module is trained on a source system (e.g., a system that is similar to the target system or a simulator), from which abundant data has been collected. The collected data can often be compactly represented by parametric regression techniques~\cite{li-icra17,zhou-cdc17}. For the source system~\eqref{eqn:system_sim_source}, we adopt the DNN module from~\cite{zhou-cdc17} and transfer this inverse module to enhance the target system~\eqref{eqn:system_sim_target} with the proposed online learning approach. The DNN module of the source system is a 3-layer feedforward network with 20 hyperbolic tangent neurons in each hidden layer. The input and output of the DNN module are $\mathcal{I}=[x(k),y_d(k+1)]$ and $\mathcal{O}=u_1(k)$. The training dataset is constructed from the source system's response on 25 sinusoidal trajectories with different combinations of frequencies and amplitudes; Matlab's Neural Network Toolbox is used to train the DNN model parameters offline. More details of the DNN training can be found in~\cite{zhou-cdc17}.

\subsubsection{Error Prediction with Online Learning}
\label{subsubsec:sim_onlineLearningModule}
The online error prediction module is a local model trained on a small dataset constructed from the latest observations of the target system. 
% (e.g., for the simulation, data from the latest 15 time steps are used). 
The objective of incorporating the online module is to achieve fast adaptation to the dynamic differences between the source and target systems. In the simulation, a Gaussian process (GP) regression model is utilized for learning the error prediction module online. 
Based on Remark~1, the input and output of the online module are selected to be $\mathcal{I}=[x(k), u_1(k), y_d(k+1)]$ and $\mathcal{O}=e_p(k+1)$, respectively.  At each time step $k$, a fixed-sized training dataset is constructed based on the latest 15 observations; in particular, the input and output are $\{(x(p-1), u(p-1), y_d(p))\}$ and $\{y_d(p) - y(p)\}$ for $p=k-15,...,k$. For the simulation, the GP model uses the squared-exponential kernel $K(\xi,\xi')=\sigma_1^2\exp\left(-\frac{1}{2}\sum_i\frac{(\xi_i-\xi_i')^2}{l_i^2}\right)$ and polynomial explicit basis functions $\{1,\xi_i,\xi_i^2\}$, where $\xi$ denotes the input to the module and $\xi_i$ denotes the $i$-th component of $\xi$, $l_i$ is the length scale associated with the input dimension $\xi_i$, and $\sigma_1^2$ is the prior variance~\cite{rasmussen2006gaussian}. 
The  length scales $l_i$ are identical for all input dimensions in the simulation; the hyperparameters of the kernel function and the coefficients of the basis functions are optimized online with Matlab's Gaussian Process Regression toolbox. The gain $\alpha^*$ is estimated based on the online error prediction module as $\hat{\alpha}^*=
-\left(\partial F_\text{gpr}/\partial u_1\right)^{-1}$, where $F_\text{gpr}$ denotes the function represented by the GP regression model.

\subsection{Simulation Results}
\label{subsec:simulationResults}
\begin{figure}[!t]
\centering
			\includegraphics[width=\columnwidth]{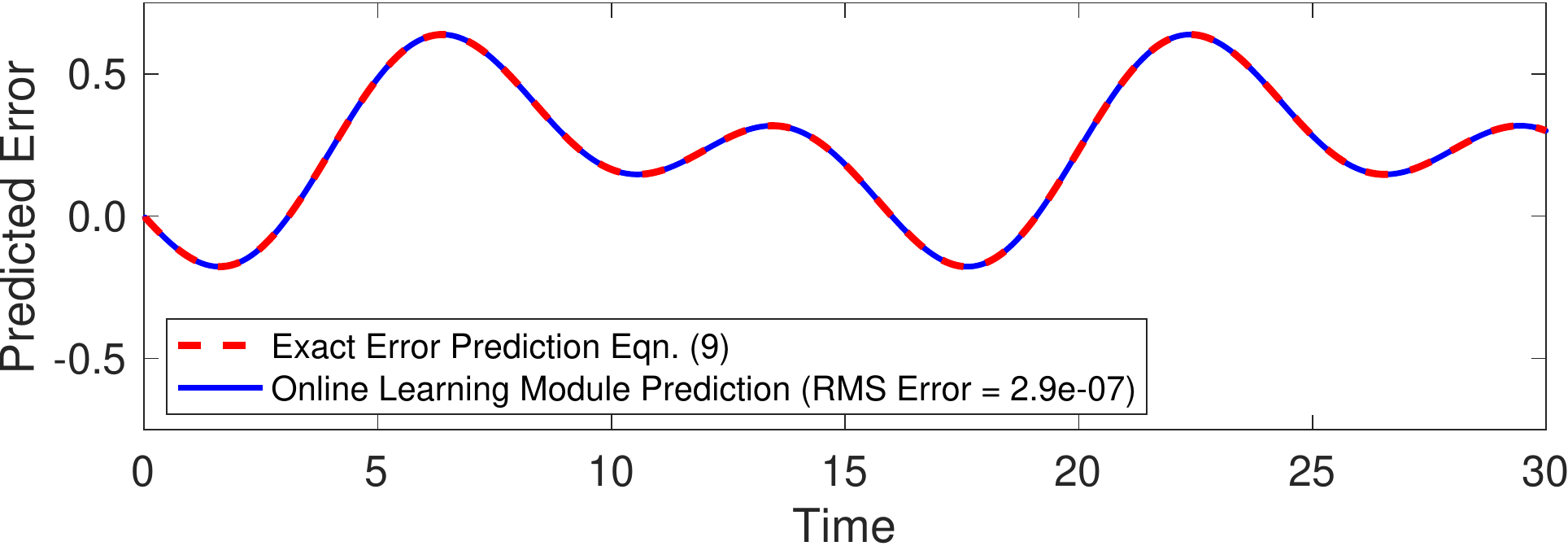}
\caption{A plot of the error prediction from the online learning module. The error predicted by an online module trained based on Remark~1 (blue) coincides with the exact error prediction computed based on Eqn.~\ref{eqn:optimalErrorPrediction} (red).}
\label{fig:sim_predictedErrorAndReference}
\end{figure}

\begin{figure}[!t]
\centering
			\includegraphics[width=\columnwidth]{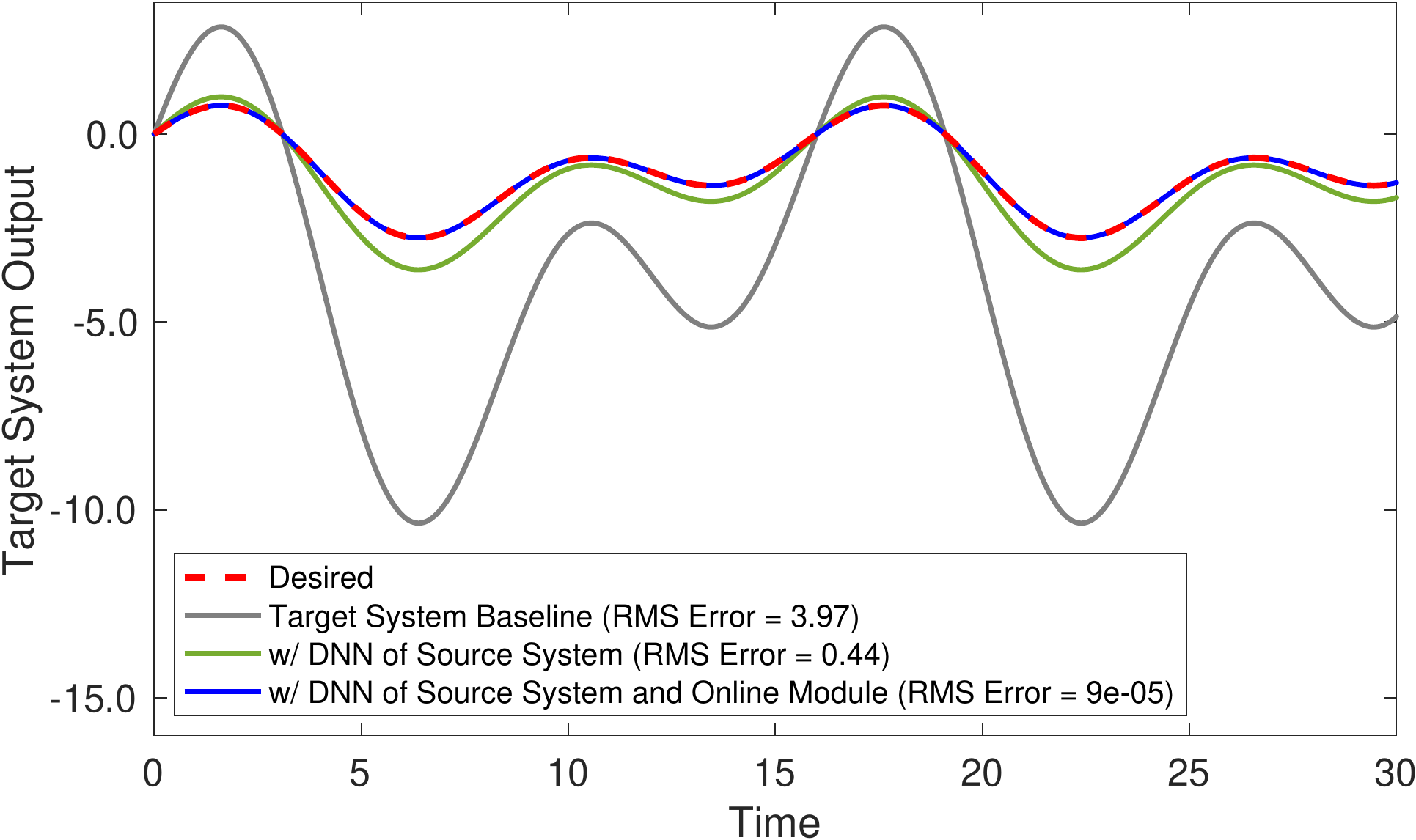}
\caption{Plots of the target system output when \textit{(i)} the baseline controller (grey), \textit{(ii)} the baseline controller with the offline learning module (green), and \textit{(iii)} the baseline controller with both the offline and online learning modules (blue) are applied. Due to system similarity, the offline learning module (trained on the source system) significantly reduces the tracking error of the target system. With the further incorporation of the online learning module, exact tracking is approximately achieved.}
\label{fig:sim_targetSystemOutput}
\vspace{-1.5em}
\end{figure}

Figures~\ref{fig:sim_predictedErrorAndReference} and \ref{fig:sim_targetSystemOutput} show the performance of the learning modules and the target system on a test trajectory $y_d(t) = \sin\left(\frac{2\pi}{8}t\right) + \cos\left(\frac{2\pi}{16}t\right) - 1$, where $t=1.5\times 10^{-3}k$ is the continuous-time variable. This test trajectory is not previously used in the training of the offline learning module. 

Figure~\ref{fig:sim_predictedErrorAndReference} compares the predicted error from the online module and the analytical error prediction of the target system computed based Eqn.~\eqref{eqn:optimalErrorPrediction}. It can be seen that the online module designed based on Remark~1 is able to accurately predict the error of the target system that would result from applying the reference $u_1$ alone. On the test trajectory, the root-mean-square (RMS) error of the online module prediction is approximately $2.9\times 10^{-7}$. 

Figure~\ref{fig:sim_targetSystemOutput} shows the outputs of the target system when \textit{(i)} the baseline controller is applied (grey), \textit{(ii)} the baseline system is enhanced by the offline module alone (green), and \textit{(iii)} the baseline system is enhanced by both the online and the offline modules (blue). As compared to the baseline system, the offline module alone reduces the RMS tracking error of the target system from $3.97$ to $0.44$. The online module further reduces the RMS tracking error to $9\times 10^{-5}$. Applying the offline and the online learning modules jointly allows the target system to achieve approximately exact tracking on a test trajectory that is not seen by the source or the target system a-priori.
\section{Quadrotor Experiments}
\label{sec:experiments}
With impromptu tracking of hand-drawn trajectories as the benchmark problem~\cite{li-icra17}, we illustrate the proposed online learning approach for transferring the DNN module trained on a source quadrotor system, the Parrot ARDrone 2.0, to a target quadrotor system, the Parrot Bebop~2. A demo video can be found here: \href{http://tiny.cc/dnnTransfer}{http://tiny.cc/dnnTransfer}

\subsection{Experiment Setup}
\label{subsec:experimentSetup}
In~\cite{li-icra17} and \cite{zhou-cdc17}, with the ARDrone as the testing platform, it is shown that a DNN module trained offline can effectively enhance the impromptu tracking performance of the quadrotor on arbitrary hand-drawn trajectories. In this work, we leverage the DNN module trained on the ARDrone to enhance the impromptu tracking performance of the Bebop and further apply the proposed online learning approach (Remark~1) to achieve high-accuracy tracking. 

\subsubsection{Control Objective and Baseline Control System} The dynamics of a quadrotor vehicle can be characterized by 12 states: translational positions $\mathbf{p}=\left(x,y,z\right)$, translational velocities $\mathbf{v}=\left(\dot{x},\dot{y},\dot{z}\right)$, roll-pitch-yaw angles $\boldsymbol{\theta}=\left(\phi,\theta,\psi\right)$, and rotational velocities $\boldsymbol{\omega}=\left(p,q,r\right)$. The objective is to design a control system such that the position of the quadrotor $\mathbf{p}_a$ tracks desired trajectories $\mathbf{p}_d$ generated from arbitrary hand drawings. In this work, we use the RMS error as the measure for evaluating tracking performance. 

The baseline control systems of the quadrotor platforms, the ARDrone and the Bebop, have an offboard position controller running at 70~Hz and an onboard attitude controller running at 200~Hz. The offboard position controller receives the reference position $\mathbf{p}_r$ and reference velocity $\mathbf{v}_r$, and computes attitude commands $\phi_\text{cmd}$ and $\theta_\text{cmd}$, yaw rate command $r_\text{cmd}$, and $z$-velocity command $\dot{z}_\text{cmd}$. The onboard attitude controller receives the commands from the offboard controller, and computes the desired thrusts of the four motors of the vehicle. In the experiments, we apply the offline and online learning modules to enhance the tracking performance of the baseline controller of the Bebop (the target system). In the design of the learning modules, we assume that the high-level dynamics of the ARDrone and the Bebop are decoupled in the $x$, $y$, and $z$ directions.


\subsubsection{DNN Module Trained on ARDrone (Source System)} In~\cite{li-icra17,zhou-cdc17}, a DNN module is trained offline to approximate the inverse of the ARDrone baseline system dynamics. Based on the theoretical insights in~\cite{zhou-cdc17}, the input and output of the DNN module are determined to be $\mathcal{I}_1=[x_d(k+4)-x_a(k),y_d(k+4)-y_a(k),z_d(k+3)-z_d(k),\dot{x}_d(k+3)-\dot{x}_a(k),\dot{y}_d(k+3)-\dot{y}_a(k),\dot{z}_d(k+2)-\dot{z}_a(k),\boldsymbol{\theta}_a(k),\boldsymbol{\omega}_a(k)]$ and $\mathcal{O}_1=[\mathbf{p}_r(k)-\mathbf{p}_a(k), \mathbf{v}_r(k)-\mathbf{v}_a(k)]$. The DNN module consists of fully-connected feedforward networks with 4 hidden layers of 128 rectified linear units (ReLU). The training dataset of the DNN module is constructed from the ARDrone baseline system response on a 400-second, 3-dimensional sinusoidal trajectory. At a sampling rate of 7~Hz, approximately 2,800 pairs of data points are collected for training. The DNN module is implemented using Tensorflow in Python. %, and the Adam optimizer is used for tuning the weight parameters of the DNN module offline~\cite{kingma2014adam}. 
Further details of the DNN module implementation can be found in~\cite{li-icra17,zhou-cdc17}. As shown in~\cite{li-icra17}, for 30 hand-drawn test trajectories, this offline DNN module is able to reduce the impromptu tracking error of the ARDrone baseline system by 43\% on average.

\subsubsection{Online Learning for Bebop (Target System)} 
%In the experiment, we adopt the ARDrone DNN module from~\cite{zhou-cdc17}, and verify the proposed online learning approach by transferring the ARDrone DNN module to the Bebop and evaluating the performance of the Bebop in impromptu tracking tasks. 
Based on Remark~1, the input and output of the online learning module are $\mathcal{I}_2=[\mathbf{p}_a(k),\mathbf{v}_a(k),\boldsymbol{\theta}_a(k),\boldsymbol{\omega}_a(k),\mathbf{p}_{r}(k),\mathbf{v}_{r}(k),x_d(k+4),y_d(k+4),z_d(k+3),\dot{x}_d(k+3),\dot{y}_d(k+3),\dot{z}_d(k+2)]$ and $\mathcal{O}_2=[x_e(k+4),y_e(k+4),z_e(k+3),\dot{x}_e(k+3),\dot{y}_e(k+3),\dot{z}_e(k+2)]$, where $(\cdot)_e$ denotes the predicted position and velocity tracking errors of the Bebop system when the offline DNN trained on the ARDrone system is used. In the experiment, in order to make online learning more efficient, instead of predicting the position and velocity errors directly, we train a GP model to predict the position of the Bebop $\mathbf{p}_a(k+r)=[x_a(k+4),y_a(k+4),z_a(k+3)]$ and computes the predicted error by subtracting the predicted position from future desired position $\mathbf{p}_d(k+r)-\mathbf{p}_a(k+r)$, where $\mathbf{p}_d(k+r)=[x_d(k+4),y_d(k+4),z_d(k+3)]$. The predicted position errors are used to compute the corrections for the position components; the velocity reference corrections are numerically approximated with a first-order finite difference scheme. For the experiments, the online learning module is implemented by using the GPy library in Python. We use a standard squared-exponential kernel with a fixed length scale $l$ for all input dimensions, prior variance $\sigma_1^2 $, and zero mean Gaussian measurement noise with variance $\sigma_2^2$~\cite{rasmussen2006gaussian}. At each time step $k$, the most recent 40 observations are used for constructing the training dataset. The hyperparameters of the GP model are $l=20$, $\sigma_1^2 = 1$, and $\sigma_2^2=2\times 10^{-5}$; these values are manually tuned a-priori for our experimental setup. If computational resources permit, we expect finer tuning of the hyperparameters online would lead to lower generalization errors and better tracking performance. \textcolor{black}{Due to the measurement noise in the experiment, instead of estimating the parameter $\alpha$ online, we used constant gains $\alpha = (5, 5, 0.5)$ for the $x$, $y$, and $z$ directions.}

\subsection{Experiment Results}
\label{subsec:experimentResults}
\begin{figure}
    \centering
     \begin{subfigure}[t]{\columnwidth}
        \centering
			\includegraphics[width=\textwidth]{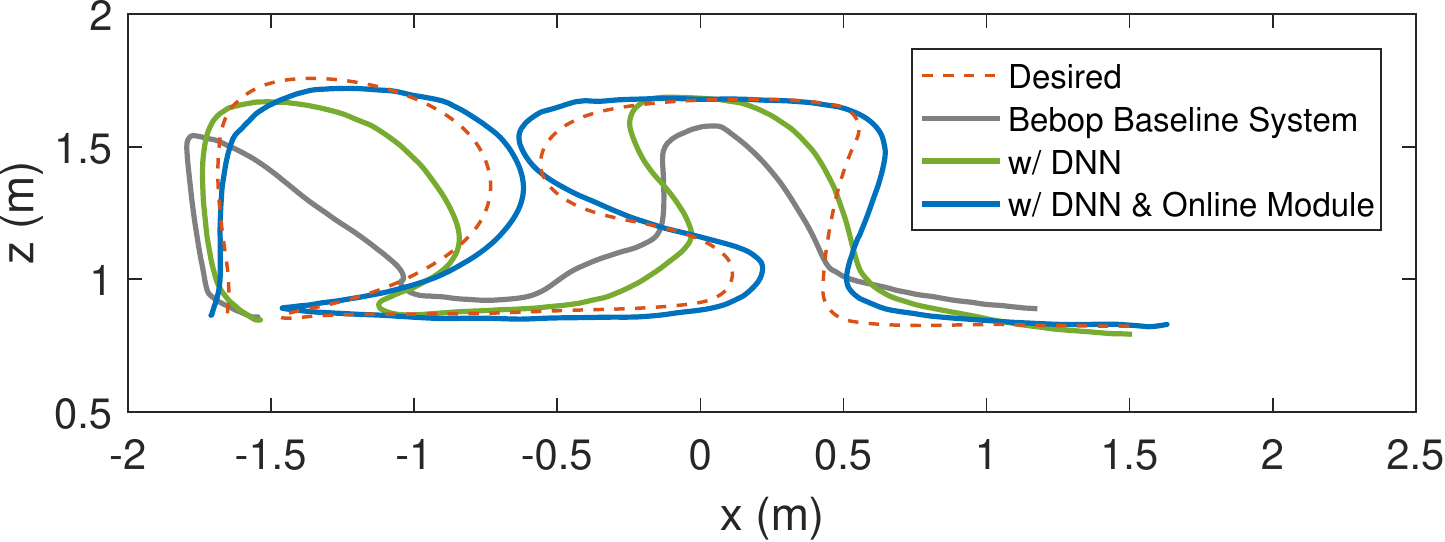}
			\vspace*{-1.5em}
        \caption{Path of the target system in the $x$-$z$ plane.}
        \label{subfig:exp_path}
    \end{subfigure}
    \noindent\\[0.5em]
    \begin{subfigure}[t]{\columnwidth}
        \centering
			\includegraphics[width=\textwidth]{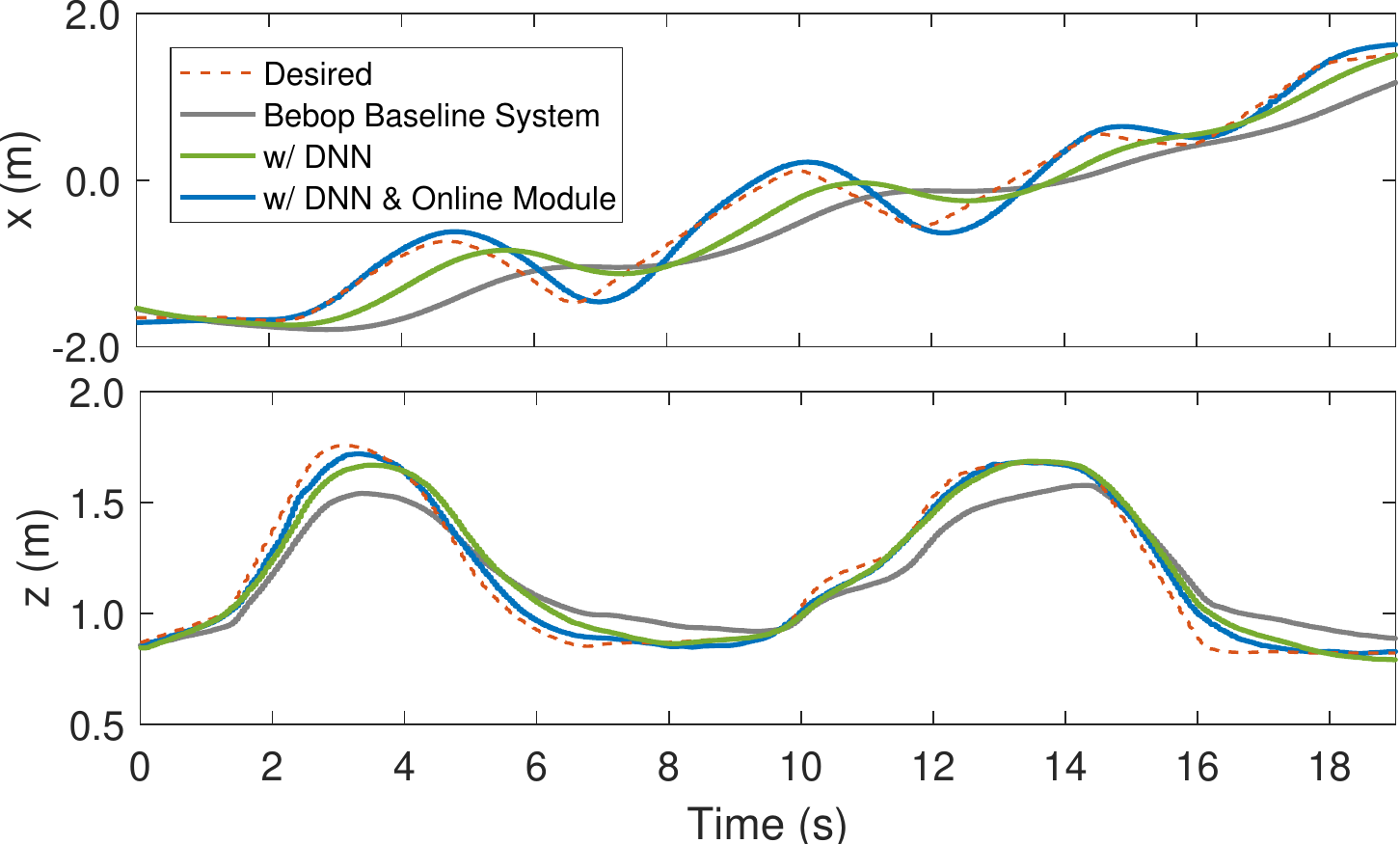}
			\vspace*{-1.5em}
        \caption{Position trajectories of the target system.}
		\label{subfig:exp_traj}
    \end{subfigure}
    \caption{Comparison of three control strategies for the Bebop target system: The RMS error is 0.42 m for the Bebop baseline system (grey), 0.26 m for the baseline system enhanced by the ARDrone DNN (green), and 0.14 m for the baseline system further enhanced by the online learning module~(blue).}
    \label{fig:dslResults}
    \vspace{-1.8em}
\end{figure}

Figure~\ref{fig:dslResults} compares the tracking performance of three control strategies on the Bebop on one of the test hand-drawn trajectories. When comparing the performance of the Bebop system enhanced by the ARDrone DNN (green) and the performance of the Bebop baseline system (grey), the ARDrone DNN reduces the delay and the amplitude errors in the Bebop tracking response. Along this particular trajectory, the DNN module alone reduces the RMS tracking error of the Bebop from approximately 0.42~m to 0.26~m. When further comparing with the performance of the DNN-enhanced system with the addition of the online learning module (blue), the tracking of the Bebop, especially in the $x$-direction, is brought close to the desired trajectory. With the online learning module, the RMS tracking error is reduced to approximately 0.14~m. Note that, from the plots in Fig.~\ref{fig:dslResults}, when the online learning module is applied, there are small overshoots at the locations with larger curvatures. The overshoots may be reduced with  online tuning of the GP hyperparameters and online estimations of the $\alpha$ parameters.

\begin{figure*}[!t]
    \centering
    \includegraphics[trim={0cm -0cm 0cm -0cm},clip, width=\textwidth]{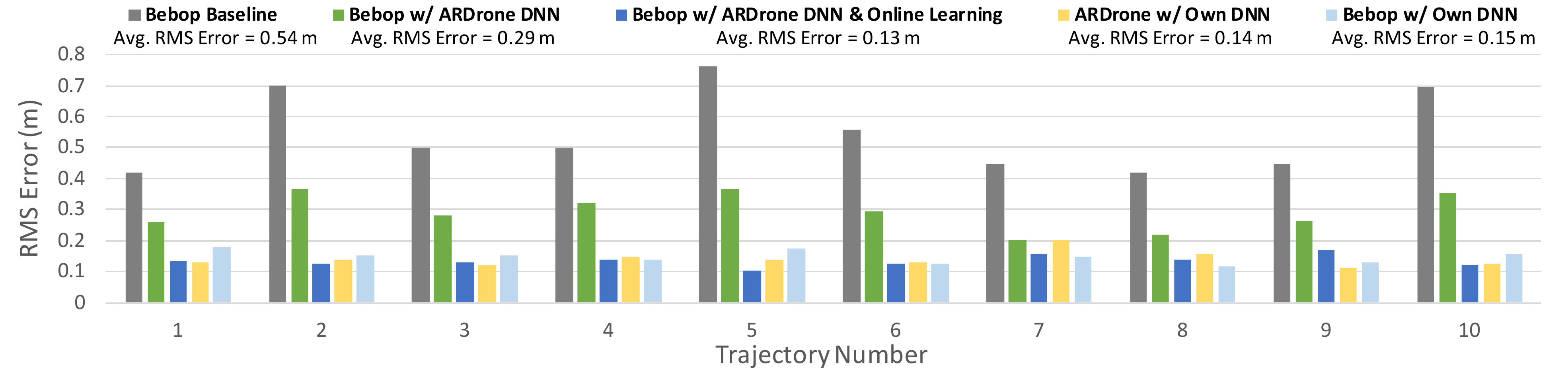}
    \caption{Tracking performance of the target system (Bebop) on 10 hand-drawn trajectories. The ARDrone DNN module alone (green) and the ARDrone DNN module with the online learning module (blue) reduce the tracking error of the Bebop baseline system (grey) by 46\% and 74\% on average, respectively. With the proposed online learning approach, the average RMS error of the Bebop (blue) is comparable to cases where the ARDrone and the Bebop are enhanced by their own offline DNN modules (yellow and light blue).
    }    
    \label{fig:RMSsummary}
    \vspace{-1em}
\end{figure*}

Figure~\ref{fig:RMSsummary} summarizes the performance errors of the three control strategies on 10 hand-drawn trajectories. When compared with the Bebop baseline system performance (grey), the direct application of the transferred DNN module (green) reduces the RMS tracking error of the Bebop baseline system by an average of 46\%. With the addition of the online learning module (blue), an average of 74\% RMS tracking error reduction is achieved. Two additional sets of results are included for comparison: \textit{(i)} the performance of the ARDrone enhanced by the DNN module trained on the ARDrone system (yellow) and \textit{(ii)} the performance of the Bebop enhanced by a DNN module trained on the Bebop system (light blue). Without requiring further data collection and offline training, the inclusion of the online learning module effectively reduces the RMS tracking error of the Bebop to values that are comparable to those of the cases where the quadrotors are enhanced by their own offline DNN modules. These results demonstrate the efficiency of the proposed online learning module to leverage past experience and reduce data re-collection and training. 
\section{Conclusions and Future Work}
\label{sec:conclusions}
In this paper, we consider the impromptu tracking problem and propose an online learning approach to efficiently transfer a DNN module trained on a source robot system to a target robot system. In the theoretical analysis, we derive an expression of the online module for achieving exact tracking. Then, based on a linear system formulation, we propose an approach for characterizing system similarity and provide insights on the impact of the system similarity on the stability of the overall system in the knowledge transfer problem. We verify our approach experimentally by applying the proposed online learning approach to transfer a DNN inverse dynamics module across two quadrotor platforms (Parrot ARDrone and Bebop). On 10 arbitrary hand-drawn trajectories, the DNN module of the source system reduces the tracking error of the target system by an average of 46\%. The incorporation of the online module further reduces the tracking error and leads to an average of 74\% error reduction. These experimental results show that the proposed online learning and knowledge transfer approach can efficaciously circumvent data recollection on the target robot, and thus, the costs and risks associated with training new robots to achieve higher performance in impromptu tasks.

Potential future work includes extending the theoretical formulation to multi-input-multi-output (MIMO) systems, extending the stability analysis to nonlinear systems, and testing the approach in different outdoor conditions and on different robot platforms.

\bibliographystyle{IEEEtran}
\bibliography{IEEEabrv,reference}

\end{document}